\documentclass[letterpaper, 10 pt, conference]{ieeeconf}  

\IEEEoverridecommandlockouts                              

\makeatletter
\let\NAT@parse\undefined
\makeatother
\usepackage[pagebackref=false,breaklinks=true,colorlinks=true,bookmarks=false,linkcolor={red!50!black},urlcolor={blue},citecolor={green!60!black}]{hyperref} 
\usepackage{url}
\usepackage{pifont}
\usepackage{booktabs} 

\usepackage{etoolbox}
\makeatletter
\patchcmd{\@makecaption}
  {\scshape}
  {}
  {}
  {}
\makeatletter
\patchcmd{\@makecaption}
  {\\}
  {.\ }
  {}
  {}
\makeatother

\usepackage{graphicx}
\usepackage{subfigure}
\usepackage{lipsum}
\usepackage{cite}
\usepackage{bm}
\usepackage{esvect}
\usepackage{amsmath}
\usepackage[ruled,vlined]{algorithm2e}
\usepackage{color}
\usepackage{amsmath,amssymb}
\usepackage{tabularx}
\usepackage{multirow}
\usepackage{arydshln}
\usepackage[dvipsnames]{xcolor}
\usepackage[table]{xcolor}

\usepackage{overpic}
\usepackage{overpic} 
\usepackage{color}
\usepackage{mathtools} 
\usepackage{currfile}
\usepackage{multirow}

\definecolor{turquoise}{cmyk}{0.65,0,0.1,0.3}
\definecolor{purple}{rgb}{0.65,0,0.65}
\definecolor{dark_green}{rgb}{0, 0.5, 0}
\definecolor{orange}{rgb}{0.8, 0.6, 0.2}
\definecolor{red}{rgb}{0.8, 0.2, 0.2}
\definecolor{darkred}{rgb}{0.6, 0.1, 0.05}
\definecolor{blueish}{rgb}{0.0, 0.3, .6}
\definecolor{light_gray}{rgb}{0.7, 0.7, .7}
\definecolor{pink}{rgb}{1, 0, 1}
\definecolor{greyblue}{rgb}{0.25, 0.25, 1}

\renewcommand{\paragraph}[1]{\vspace{.5em}\noindent\textbf{#1}.}

\usepackage{lipsum}

\usepackage{blindtext}

\newcommand{\kostas}[1]{{\color{Bittersweet} {[\bf Kosta: #1]}}}
\newcommand{\andrew}[1]{{\color{BlueViolet} {[Andrew: #1]}}}
\newcommand{\vitto}[1]{{\color{OrangeRed} {[Vitto: #1]}}}
\newcommand{\JB}[1]{{\color{OliveGreen} {[Jon: #1]}}}
\newcommand{\tom}[1]{{\color{RoyalPurple} {[Tom: #1]}}}
\newcommand{\pratul}[1]{{\color{Emerald} {[Pratul: #1]}}}
\newcommand{\at}[1]{{\color{blueish}#1}}
\newcommand{\AT}[1]{{\color{blueish}{\bf [Andrea: #1]}}}
\newcommand{\At}[1]{\marginpar{\tiny{\textcolor{blueish}{#1}}}}
\newcommand{\al}[1]{\textbf{\color{orange}[AL: #1]}}
\renewcommand{\kostas}[1]{}
\renewcommand{\andrew}[1]{}
\renewcommand{\vitto}[1]{}
\renewcommand{\JB}[1]{}
\renewcommand{\tom}[1]{}
\renewcommand{\pratul}[1]{}
\renewcommand{\at}[1]{}
\renewcommand{\AT}[1]{}
\renewcommand{\At}[1]{}
\renewcommand{\al}[1]{}

\makeatletter
\usepackage{xspace}
\DeclareRobustCommand\onedot{\futurelet\@let@token\@onedot}
\def\@onedot{\ifx\@let@token.\else.\null\fi\xspace}

\makeatother

\definecolor{gold}{rgb}{0.82, 0.53, 0.04}
\definecolor{silver}{rgb}{0.43, 0.47, 0.35}
\definecolor{bronze}{rgb}{0.5, 0.51, 0.78}

\title{\LARGE \bf

Pocket-SLAM: Rendering-Area-Aware Pruning for Memory-Efficient 3DGS-SLAM
}

\author{
Leshu Li$^{1*}$,
\thanks{$^{1}$University of Minnesota, Twin Cities, USA. 
{\tt\small li00385@umn.edu}}%
\thanks{$^{2}$University of North Carolina at Chapel Hill, USA.}%
\thanks{*Corresponding author}
Jie Peng$^{2}$,
Yang Zhao$^{1}$
}

\begin{document}

\maketitle
\thispagestyle{empty}
\pagestyle{empty}

\begin{abstract}
3D Gaussian Splatting (3DGS) has garnered significant attention in Simultaneous Localization and Mapping (SLAM) due to its advances in capturing fine-grained geometry features and synthesizing novel views.
For SLAM in large-scale scenes, such as autonomous driving, 3DGS-SLAM faces a critical limitation. The memory consumption increases continuously over time as Gaussian points accumulate, leading to poor memory efficiency and limiting its applicability.
In this work, we propose a rendering-area–aware pruning strategy that selectively removes Gaussians based on their contribution to the effective rendering area, rather than solely relying on Gaussian-level heuristics (e.g., opacity or gradient magnitude). This perspective directly targets the sources of memory redundancy, effectively reducing the peak memory footprint of 3DGS-SLAM during runtime. Evaluations on the EuRoC and KITTI datasets demonstrate that our method consistently outperforms existing pruning approaches in large-scale outdoor scenes, achieving over 60\% memory reduction and more than $2\times$ FPS improvement while preserving localization and mapping accuracy. These results highlight rendering-area–aware pruning as a promising direction for scaling 3DGS-SLAM to real-world autonomous driving scenarios. Our code is publicly available at \url{https://github.com/UMN-ZhaoLab/Pocket-SLAM.git}.

\end{abstract}

\section{Introduction}

Visual SLAM is fundamental to robot perception and underpins a wide range of applications, including autonomous navigation, environment mapping, and interactive perception. 3D Gaussian Splatting (3DGS)~\cite{kerbl3Dgaussians} has recently attracted significant attention in SLAM as it enables fine-grained geometric representation and high-quality novel view synthesis. For instance, MonoGS~\cite{MonoGS} and GS-SLAM~\cite{GSSLAM} demonstrate that 3DGS-SLAM can achieve both high-fidelity scene reconstruction and accurate camera tracking, highlighting its strong potential for applications such as autonomous driving~\cite{drive1,drive2} and drone navigation~\cite{drone1}. Moreover, LGS-SLAM~\cite{LGSSLAM} and WildGS-SLAM~\cite{wildslam} extend 3DGS-SLAM to large-scale outdoor scenarios, validating its feasibility in real-world driving environments. Nevertheless, a critical limitation persists: large-scale deployments introduce substantial memory redundancy. While the number of Gaussians in compact indoor scenes is relatively modest, vast and unstructured outdoor environments necessitate storing and updating millions of Gaussians in real time, resulting in prohibitively high peak memory consumption that becomes a major bottleneck for deployment. This challenge is further exacerbated by the increasing demand for 3DGS-SLAM to operate efficiently on resource-constrained edge devices (e.g., GPUs embedded in autonomous vehicles or drones). In this context, reducing peak runtime memory consumption emerges as a central requirement for practical deployment.

Existing works have investigated Gaussian pruning in the context of 3DGS. However, most of these efforts focus primarily on small-scale indoor scenarios~\cite{pruning1,pruning2,pruning3,pruning4} or address only the storage of 3DGS-SLAM keyframes~\cite{GEVO}, while neglecting the critical issue of peak runtime memory consumption. This limitation is particularly important, as peak memory directly determines the feasibility of deploying 3DGS-SLAM on edge devices with limited memory capacity, and therefore has far greater practical implications for real-world large-scale applications. Furthermore, indoor and outdoor scenarios exhibit fundamentally different characteristics: while indoor environments are relatively compact and texture-rich, outdoor environments often consist of vast regions with sparse or low-texture areas. These differences result in a significantly larger number of Gaussians being generated and continuously updated during outdoor mapping, which further exacerbates the challenge of peak runtime memory consumption.

\begin{figure}[t]
    \centering
    \includegraphics[width=1\linewidth]{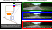}
    \caption{Results on KITTI~\cite{KITTI} sequence 10. Compared with LSG-SLAM~\cite{LGSSLAM}, our method achieves comparable camera tracking accuracy (Absolute Trajectory Error Root Mean Square Error, ATE) and rendering quality (Peak Signal-to-Noise Ratio, PSNR) in large-scale scenarios, while reducing peak memory usage by 61\%.}
    \label{fig:intro}
    \vspace{-5mm}
\end{figure}

In this paper, we present Pocket-SLAM, a practical and seamlessly integrated extension to 3DGS-SLAM that incorporates a \textit{rendering-area–aware pruning strategy}. During mapping, our method quantifies the contribution of each Gaussian by computing its effective pixel coverage on the image plane and prunes those with negligible rendering impact. Rendering-area–aware pruning measures Gaussian importance from the perspective of their contribution to the final rendered image, ensuring that in large-scale scenes, Gaussians covering critical regions are preferentially preserved. However, applying rendering-area–aware pruning alone, while effective in reducing memory, inevitably results in severe information loss in texture-dense regions. To mitigate this issue, we introduce a \textit{tile-level budget mechanism}, which adaptively constrains the pruning ratio within each tile to prevent over-pruning in both texture-dense and texture-scarce areas, thereby maintaining balanced Gaussian distributions and preserving texture information. As illustrated in Fig.~\ref{fig:intro}, our method achieves substantial memory savings in outdoor environments while maintaining both tracking and mapping accuracy. Furthermore, our approach is orthogonal to other 3DGS acceleration techniques and can be readily combined with them for further performance improvements.

In summary, the contributions of our work include:

\begin{itemize}
    \item We propose a \textit{rendering-area–aware pruning strategy} that assesses Gaussian importance based on effective pixel coverage, shifting the criterion from local heuristics to scene-level rendering efficiency.
    \item We introduce a \textit{tile-level budget mechanism} that adaptively constrains pruning according to per-tile Gaussian allocation, preventing excessive pruning in both texture-dense and texture-sparse regions, and ensuring balanced Gaussian distributions with robust texture preservation.
    \item We demonstrate through evaluations on large-scale outdoor environments that our method outperforms mainstream pruning strategies, reducing peak memory consumption by over 60\% and improving FPS by more than $2\times$ while preserving localization and mapping accuracy.
\end{itemize}
\section{Related Work}
\label{sec:related_work}

\subsection{Traditional Visual SLAM}
Visual SLAM has been extensively studied in both indoor and outdoor scenarios, yet the challenges differ fundamentally. Indoor SLAM frameworks~\cite{orbslam,orbslam3,indoorslam3,indoorslam4,indoorslam5} operate in compact environments with abundant textures and structural regularities, enabling reliable tracking with sparse feature-based, direct, or dense methods. These approaches, however, are less effective in large-scale outdoor environments, which are often unstructured and contain vast texture-sparse regions. Outdoor SLAM frameworks, such as stereo-based~\cite{stereo1,stereo2} and LiDAR-based systems~\cite{Lidar1,Lidar2}, address scale and robustness but require specialized sensors and still struggle to jointly achieve efficiency and high-fidelity reconstruction.

\subsection{3DGS-SLAM in Outdoor Environments}
With the advent of 3DGS~\cite{kerbl3Dgaussians}, researchers have increasingly integrated Gaussian primitives into SLAM systems. MonoGS~\cite{MonoGS} and GS-SLAM~\cite{GSSLAM} show the feasibility of combining 3DGS with monocular SLAM for accurate tracking and high-fidelity reconstruction. Photo-SLAM~\cite{photoslam} and LoopSplat~\cite{Loopsplat} incorporate ORB-SLAM~\cite{orbslam} modules or global BA to enhance robustness, but most evaluations remain limited to indoor scenes. Recent work such as LGS-SLAM~\cite{LGSSLAM} and WildGS-SLAM~\cite{wildslam} extends 3DGS-SLAM to outdoor driving scenarios, highlighting its potential for large-scale deployment. However, outdoor environments require millions of Gaussians to represent unstructured regions, resulting in prohibitively high peak memory consumption and limiting deployment on resource-constrained edge devices such as those in autonomous vehicles and drones.

\subsection{Memory-Efficient Gaussian Pruning}
To mitigate redundancy in 3DGS models, recent research has focused on Gaussian pruning. 
LightGaussian~\cite{pruning1} prunes Gaussians based on opacity, removing those with negligible contributions to the rendered image. 
LP-3DGS~\cite{pruning3} uses gradient magnitude as an importance criterion, discarding Gaussians with small parameter updates. 
PUP-3DGS~\cite{pruning2} removes Gaussians with little perceptual utility, while MaskGaussian~\cite{pruning4} temporarily masks low-weight Gaussians before pruning them. 
Although effective for 3DGS rendering and training, these methods cannot be directly applied to 3DGS-SLAM, which demands per-frame rendering and tracking accuracy. 
GEVO~\cite{GEVO} is among the few works addressing memory in 3DGS-SLAM, but it focuses only on storage via reduced keyframe retention rather than runtime peak memory consumption. 
This limitation motivates our work: we propose Pocket-SLAM, which introduces a \textit{rendering-area--aware pruning strategy} and a \textit{tile-level budget mechanism} to reduce peak memory consumption in large-scale outdoor environments while preserving tracking accuracy and reconstruction fidelity.

\section{Proposed Methods}
\label{sec:method}

\begin{figure*}[t]
    \centering
    \includegraphics[width=0.9\linewidth]{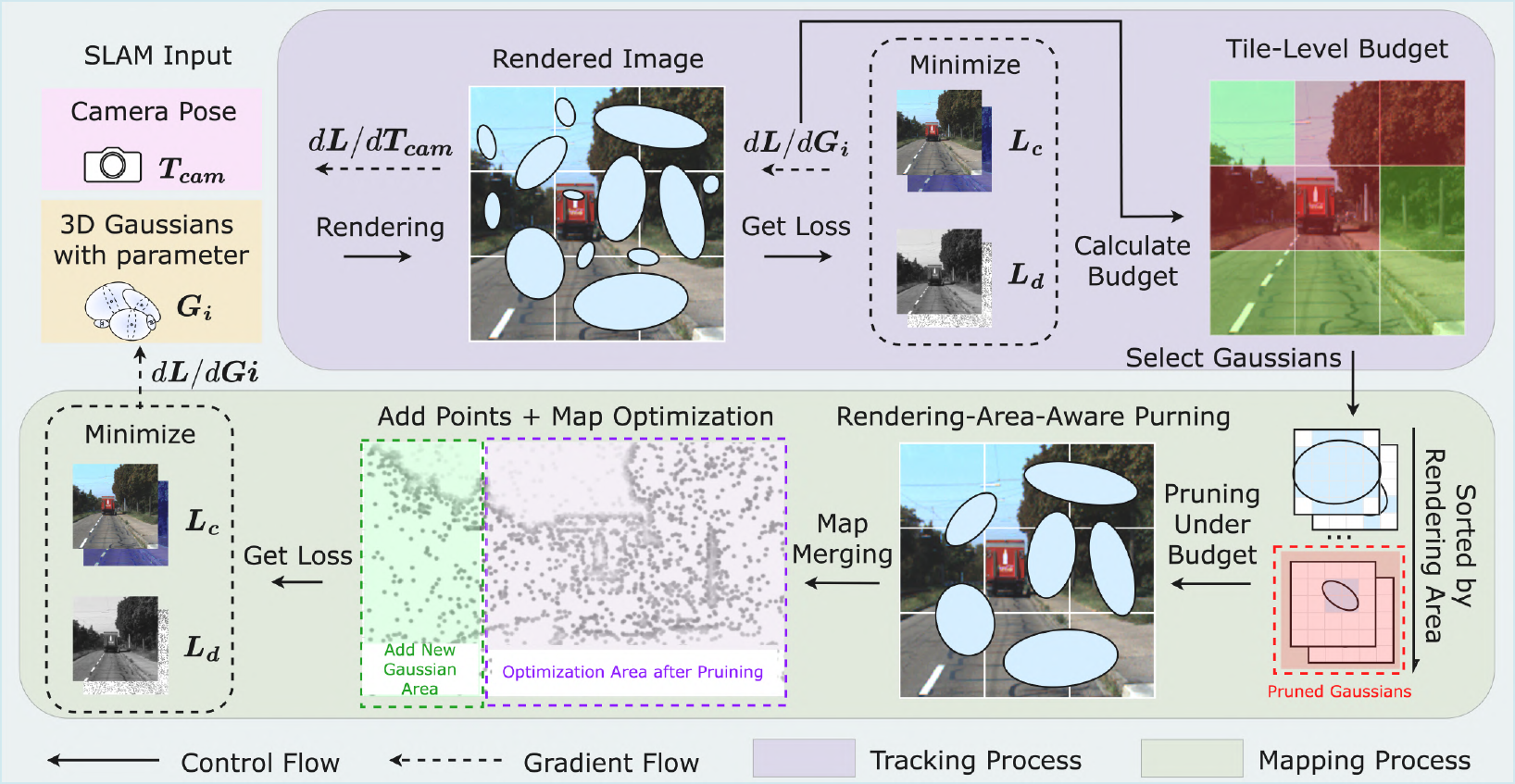}
    \caption{The overview pipeline of Pocket-SLAM. For each incoming frame, the pose is initialized with multi-modality priors and optimized using rendering and feature warping losses (tracking). Keyframes are then selected to refine the scene and insert new Gaussians (mapping). Our pruning strategies are integrated into mapping: the \emph{Tile-Level Budget Mechanism} assigns balanced survival budgets during tracking to guide pruning, and \emph{Rendering-Area--Aware Pruning} removes low-contribution Gaussians after mapping optimization, enabling memory-efficient SLAM without compromising accuracy.}
    \label{fig:pipeline}
    \vspace{-5mm}
\end{figure*}

To achieve efficient and balanced pruning, Pocket-SLAM introduces two complementary strategies: the \emph{rendering-area–aware pruning strategy} (Sec.~\ref{sec:rendering}) and the \emph{tile-level budget mechanism} (Sec.~\ref{sec:tile}). The \emph{rendering-area–aware pruning strategy} assesses the contribution of each Gaussian by explicitly computing its effective pixel coverage on the image plane. However, pruning solely based on rendering area tends to eliminate Gaussians with small coverage even in texture-dense regions, leading to significant information loss. To address this issue, we introduce the \emph{tile-level budget mechanism}, which allocates Gaussian survival budgets across different image regions. The combination of these mechanisms prevents over-pruning in texture-rich areas while still achieving substantial reductions in redundant Gaussians. Overall, this design balances SLAM accuracy with memory efficiency, making Pocket-SLAM particularly well-suited for large-scale outdoor environments.

Pocket-SLAM is built upon the standard SLAM framework, which generally comprises two fundamental stages: \emph{tracking} and \emph{mapping}. In the tracking stage, Pocket-SLAM allocates budgets to each tile based on the gradient characteristics of its Gaussians. In the mapping stage, it ranks Gaussians according to both the tile-level budgets and their rendering areas, pruning those deemed less important. The overall pipeline of Pocket-SLAM is illustrated in Fig.~\ref{fig:pipeline}.

\subsection{SLAM Pipeline}
The Pocket-SLAM framework follows the standard SLAM paradigm, alternating between \emph{tracking}, which estimates camera poses, and \emph{mapping}, which refines the Gaussian scene representation. 
We denote the camera pose as $T_{\mathrm{cam}} \in SE(3)$, which consists of a 3D rotation and translation that transform world coordinates into the camera coordinate system. 
The set of 3D Gaussians is represented as $\{G_i\}_{i=1}^N$, where each $G_i$ encodes position, covariance, opacity, and color parameters. 
For an incoming frame, the rendered color and depth at pixel $\mathbf{p}$ are denoted by 
$C(\mathbf{p}; G_i, T_{\mathrm{cam}})$ and 
$Z(\mathbf{p}; G_i, T_{\mathrm{cam}})$, respectively. 
We then define the photometric loss $L_c$ and depth loss $L_d$ as:

\begin{align}
L_c &= \sum_{\mathbf p\in\Omega_t}
\rho\!\Big(I_t(\mathbf p)-C_t(\mathbf p;G_i,T_{\mathrm{cam}})\Big)^2, \label{eq:Lc}\\
L_d &= \sum_{\mathbf p\in\Omega_t}
\psi\!\Big(D_t(\mathbf p)-Z_t(\mathbf p;G_i,T_{\mathrm{cam}})\Big)^2, \label{eq:Ld}
\end{align}
where $I_t$ and $D_t$ denote the observed color and depth, and $\rho,\psi$ are robust penalty functions. $\Omega_t$ represents the set of all pixel coordinates in the image domain of frame $t$.

\paragraph{Tracking}
In the tracking stage, the Gaussian set $G_i$ is fixed, and only the current estimated camera pose $T_{\mathrm{cam}}$ is optimized by minimizing:

\begin{equation}
\label{eq:trk_total}
L = L_c + \lambda_d\, L_d,
\end{equation}
where $\lambda_d \geq 0$ balances the two losses.  
During backpropagation, gradients are first propagated to each Gaussian and then to the camera pose:
\begin{align}
\frac{\partial L_{\mathrm{}}}{\partial G_i}
\;\longrightarrow\;
\frac{\partial L_{\mathrm{}}}{\partial T_{\mathrm{cam}}}
\;\longrightarrow\;
T_{\mathrm{cam}}\ \text{update}.
\end{align}
The per-frame gradient magnitude of each Gaussian is:
\begin{equation}
\label{eq:gi_track}
g_i = \big\|\nabla_{G_i} L\big\|_2,
\end{equation}
which is not used to update $G_i$ during tracking but is later reused to compute tile-level Gaussian budgets.  
The pose update is performed in Lie algebra form:
\begin{equation}
T_{\mathrm{cam}} \leftarrow \exp\!\big(\widehat{\delta\boldsymbol\xi}\big)\, T_{\mathrm{cam}}, 
\quad \delta\boldsymbol\xi = -\eta\,\nabla_{T_{\mathrm{cam}}} L,
\end{equation}
where $\eta$ is the learning rate and $\widehat{(\cdot)}$ is the $\mathfrak{se}(3)$ hat operator.

\paragraph{Mapping}
In the mapping stage, the camera poses $T_{\mathrm{cam}}$ are fixed, and the Gaussian set $\{G_i\}$ is optimized by minimizing
\begin{equation}
\label{eq:map_total}
L =
L_c + \lambda_d^\star\, L_d.
\end{equation}
Here, $\lambda_d^\star$ denotes a potentially different weighting coefficient used in mapping to balance photometric and depth terms.
Gradients are propagated to each Gaussian parameter and used for updates:
\begin{align}
\nabla_{G_i} L &= \nabla_{G_i}\!\big(L_c + \lambda_d L_d\big), \\
G_i &\leftarrow G_i - \alpha\, \nabla_{G_i} L,
\end{align}
where $\alpha$ is the learning rate.  
The mapping optimization continues until convergence, after which the rendering-area--aware pruning is performed to remove redundant Gaussians.

\subsection{Rendering-Area--Aware Pruning}
\label{sec:rendering}
After the mapping stage has converged, we prune Gaussians based on their rendering contribution in the current frame. 
For each Gaussian $G_i$, the pixel-level contribution is:
\begin{equation}
\label{eq:alpha}
\alpha_{i}(\mathbf p)=o_i\cdot\exp\!\left(
-\tfrac12(\mathbf p-\boldsymbol u_{i})^\top\Lambda_{i}^{-1}(\mathbf p-\boldsymbol u_{i})
\right),
\end{equation}
where $\boldsymbol u_{i}$ and $\Lambda_{i}$ denote the projected mean and covariance. 
Aggregating over all pixels yields the coverage
\begin{equation}
\label{eq:cover}
C_i=\sum_{\mathbf p\in\Omega}\alpha_{i}(\mathbf p), 
\qquad 
S_i=\frac{C_i}{\sum_j C_j}.
\end{equation}

Within each tile $k$, we rank Gaussians by $S_i$ and retain only the top $B_k^{\text{trk}}$, where $B_k^{\text{trk}}$ is determined by the Tile-Level Budget Mechanism to be defined in the next subsection. 
Since newly added Gaussians in the current mapping round have not yet undergone tracking-based budget allocation, they are exempt from pruning until subsequent iterations. 

In outdoor SLAM scenarios, particularly in autonomous driving datasets, Gaussians with large rendering areas frequently correspond to critical structures such as roads and sky.
These regions are essential for robust localization and navigation, making Gaussians with larger rendering areas especially important in outdoor environments. 
However, relying solely on rendering contribution introduces several challenges in complex outdoor scenes, which motivates our Tile-Level Budget Mechanism(Sec.~\ref{sec:tile}).

\subsection{Tile-Level Budget Mechanism}
\label{sec:tile}
Outdoor SLAM scenes are often highly complex~\cite{outdoorslam1, outdoorslam2, outdoorslam3}: some regions are texture-rich and information-dense, while others are relatively sparse.  
If pruning were conducted solely based on rendering area, two key issues would emerge.  
First, texture-rich regions are typically represented by a large number of small Gaussians.  
Uniform pruning by area would risk eliminating nearly all Gaussians in these regions, leaving them empty.  
Second, texture-sparse regions can often be represented by only a few Gaussians.  
As the global pruning ratio increases, removing these few Gaussians would result in the complete loss of information in such regions.  

\begin{figure}[t]
    \centering
    \includegraphics[width=1\linewidth]{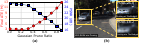}
    \caption{Results on KITTI~\cite{KITTI} sequence 3. Experiments are conducted using LSG-SLAM~\cite{LGSSLAM} as the baseline. (a) illustrates the variation of ATE and PSNR under different pruning ratios when applying Rendering-Area--Aware Pruning without the Tile-Level Budget. (b) shows a sample scene from sequence 13 at a pruning ratio of 0.6, comparing results with and without the Tile-Level Budget. This visualization highlights why increasing the pruning ratio leads to a significant degradation in SLAM accuracy.}
    \label{fig:tile}
    \vspace{-5mm}
\end{figure}

As illustrated in Fig.~\ref{fig:tile}(a), when the pruning ratio gradually increases, both tracking accuracy (ATE) and mapping accuracy (PSNR) degrade significantly.  
The root cause is that pruning based solely on rendering area tends to prioritize the removal of small Gaussians.  
However, these small Gaussians are predominantly concentrated in texture-rich regions, and deleting them without restriction eliminates most of the Gaussians in these areas.  
Consequently, the rendered images in these regions experience a substantial loss of fidelity, as shown in Fig.~\ref{fig:tile}(b), ultimately compromising overall SLAM accuracy.  
Therefore, introducing a budget mechanism to prevent regions from being entirely depleted is essential for ensuring reliable SLAM performance.

To this end, we leverage the Gaussian gradients obtained in tracking to assign a survivor budget for each image tile. 
For the set of Gaussians $\{G_i\}$ projected into tile $\mathcal{T}_k$, we compute the average gradient magnitude:
\begin{equation}
\label{eq:Gk}
G_k = \frac{1}{N_k}\sum_{i\in\mathcal T_k} g_i,
\end{equation}
where $N_k$ is the number of Gaussians in tile $k$. 
A higher $G_k$ indicates that the region is texture-dense and carries strong constraints, thus requiring a larger budget; conversely, a lower $G_k$ suggests weaker constraints. Given a global target $N_{\text{tar}}$, we allocate per-tile budgets as:
\begin{equation}
\label{eq:budget}
B_k^{\text{trk}}=\operatorname{clip}\!\left(
\Big\lceil N_{\text{tar}}\cdot \tfrac{G_k}{\sum_j G_j}\Big\rceil, 
B_{\min}, B_{\max}
\right),
\end{equation}
where $B_{\min}$ and $B_{\max}$ ensure that no tile is completely depleted or overly concentrated. 

This process is performed during the tracking stage. 
Since $G_i$ are not updated in tracking, the resulting tile-level budgets remain relatively stable across iterations, providing consistent and reliable guidance for subsequent rendering-area--aware pruning. 
In particular, texture-rich regions with larger gradients are allocated higher budgets, whereas texture-sparse regions are assigned lower budgets. 
The key principle of our design is to preserve information across all regions, ensuring that neither dense nor sparse areas suffer complete information loss. 
By combining this budget allocation with rendering-area--aware pruning, our method effectively balances memory efficiency and reconstruction fidelity.

\section{Experimental Results}
\label{sec:experiment}

In this section, we present extensive real-world experiments to evaluate the effectiveness of the proposed Pocket-SLAM. 
We further conduct systematic comparisons with state-of-the-art 3DGS pruning algorithms, and the results demonstrate that in complex large-scale outdoor scenarios, Pocket-SLAM delivers superior performance and higher efficiency compared to existing methods.

\textbf{Datasets}. We conduct evaluations on two widely used outdoor benchmarks. 
The EuRoC MAV dataset~\cite{EUROC} includes outdoor sequences MH01–MH05, recorded by a micro aerial vehicle in challenging environments with significant viewpoint and illumination variations. 
The KITTI odometry dataset~\cite{KITTI} provides large-scale outdoor driving sequences (00–10, except 01), captured by a vehicle equipped with stereo cameras and GPS/IMU sensors, covering diverse urban, residential, and rural scenes. 
Results for sequence 01 are omitted, as all methods perform poorly on it.

\textbf{Metrics}. We evaluate Pocket-SLAM from two perspectives and compare it with competitive baselines. The first perspective is \textbf{\textit{Accuracy}}. For trajectory accuracy, we report the root-mean-square error (RMSE) of the absolute trajectory error (ATE)~\cite{ATE}. For reconstruction quality, we measure peak signal-to-noise ratio (PSNR), structural similarity index (SSIM)~\cite{SSIM}, and learned perceptual image patch similarity (LPIPS)~\cite{LPIPS}. The second perspective is \textbf{\textit{Performance}}. We use frames per second (FPS) and peak memory consumption as evaluation metrics. Notably, peak memory consumption accounts for all intermediate parameters involved in 3DGS-SLAM computation, reflecting the minimum memory required for a device to execute 3DGS-SLAM.

\textbf{Baselines}. Our proposed Pocket-SLAM is implemented on top of the LSG-SLAM~\cite{LGSSLAM} pipeline, while its pruning strategy is generic and can be seamlessly integrated into other SLAM frameworks. To validate its effectiveness, we compare it against several representative pruning-based methods in the 3DGS literature, including LightGaussian~\cite{pruning1}, which prunes Gaussians primarily based on opacity, removing those with negligible contributions to the rendered image; LP-3DGS~\cite{pruning3}, which employs gradient magnitude as an importance signal and discards Gaussians with limited updates during optimization; and MaskGaussian~\cite{pruning4}, which adopts a masking--then--pruning scheme, where Gaussians with low rendering weights are first deactivated before being permanently pruned. These pruning methods, as well as Pocket-SLAM, are applied within the 3DGS-SLAM pipeline for comparison. It is worth noting that PUP-3DGS~\cite{pruning2} is designed from the perspective of perceptual image quality, which does not directly align with tracking accuracy, and thus is not included in our comparison. Our method is orthogonal to other 3DGS acceleration techniques~\cite{AdR-Gaussian, TC-GS, SG-Splatting} and can be combined with them for further performance improvement. For fair evaluation, all methods are compared under the same pruning ratio on the NVIDIA A6000~\cite{NVIDIA_A6000} platform.

\textbf{Implementation Details}. We perform 50 iterations per frame in the tracking stage and 100 iterations per keyframe in the mapping and loop optimization stages. 
For loss weighting, we set $\lambda_d = 1.0$ in Eq.~\ref{eq:trk_total} and $\lambda_d^\star = 1.5$ in Eq.~\ref{eq:map_total}. 
For pruning, the global target Gaussian count $N_{\text{tar}}$ is set to $0.4N_{\text{init}}$, where $N_{\text{init}}$ denotes the number of Gaussians before pruning. 
The minimum per-tile budget $B_{\min}$ is fixed at 5 to prevent regions from being completely emptied, while the maximum budget $B_{\max}$ is set to 200 to avoid excessive concentration of Gaussians within a single tile. 
These hyperparameters ensure balanced pruning across dense and sparse regions while preserving reconstruction fidelity.

\subsection{Evaluation on EuRoC}
\textbf{Tracking and Mapping Accuracy.}
\label{sec:euroc-tracking}
Comprehensive evaluation results are presented in Tab.~\ref{tab:accuracy-euroc}. 
Under a unified pruning ratio, only our method achieves tracking and mapping accuracy comparable to LSG-SLAM~\cite{LGSSLAM}. 
In contrast, LightGaussian~\cite{pruning1} and LP-3DGS~\cite{pruning3} often fail on challenging sequences (e.g., MH04, MH05) because their pruning relies on local Gaussian heuristics (e.g., opacity or gradient magnitude) that do not capture each Gaussian’s image contribution, frequently removing essential Gaussians in texture-sparse regions and causing drift. 
MaskGaussian~\cite{pruning4}, while completing all five sequences, still shows a notable accuracy drop. 
It prunes by global Gaussian importance but neglects the need to preserve Gaussians critical for reconstructing the \emph{current} frame, where per-frame fidelity is fundamental for stable tracking. 
By contrast, our rendering-area--aware pruning with the tile-level budget mechanism evaluates each Gaussian’s effective pixel coverage on the current frame and ensures balanced preservation across texture-rich and texture-sparse regions, thereby maintaining stable and reliable tracking and high-fidelity reconstruction.

\begin{table}[ht]
\setlength{\tabcolsep}{0.8mm}
\renewcommand\arraystretch{1.1}
\caption{Camera tracking and rendering results on EuRoC~\cite{EUROC}. Tracking is evaluated by ATE RMSE $\downarrow$ [m], Rendering is evaluated by PSNR $\uparrow$ [dB], SSIM $\uparrow$ [0-1], and LPIPS $\downarrow$ [0-1].  "-" means tracking lost.}
\vspace{-2mm} 
\label{tab:accuracy-euroc}
\hspace*{0cm}\makebox[\linewidth][c]{
\begin{tabular}{cccccccc}
\toprule
Method                   &Metric& MH01 & MH02 & MH03 & MH04 & MH05 & Avg. \\ \hline
\multirow{4}{*}{LSG-SLAM~\cite{LGSSLAM}}
    & ATE$\downarrow$  & \textbf{0.05} & \textbf{0.02} & \textbf{0.08} & 0.06 & 0.11 & { \textcolor{brown}{0.06}}\\ 
    & PSNR$\uparrow$  & \textbf{31.23} & \textbf{33.31} & 30.26 & \textbf{30.72} & \textbf{28.54} & {\textcolor{brown}{30.81}} \\  
    & SSIM$\uparrow$  & \textbf{0.98} & \textbf{0.99} &\textbf{ 0.98} & \textbf{0.98 }& \textbf{0.96} & { \textcolor{brown}{0.98}}\\
    & LPIPS$\downarrow$ & \textbf{0.04} & \textbf{0.03} & \textbf{0.05 }& \textbf{0.05} & \textbf{0.07} & { \textcolor{brown}{0.05}}\\ \hline
\multirow{4}{*}{LightGaussian~\cite{pruning1}} 
    & ATE$\downarrow$  & 1.32 & 0.94 & - & 1.67 & - & 1.31\\
    & PSNR$\uparrow$  & 13.24 & 12.85 & - & 10.94 & - & 12.34\\  
    & SSIM$\uparrow$  & 0.54 & 0.55 & - & 0.50& - & 0.53\\  
    & LPIPS$\downarrow$ & 0.41 & 0.39 & - & 0.45 & - & 0.41\\ \hline
\multirow{4}{*}{\shortstack{LP-3DGS~\cite{pruning3}}}
    & ATE$\downarrow$  & 0.44 & 0.35 & 1.34 & 0.54 & - & 0.67\\
    & PSNR$\uparrow$  & 15.65 & 16.25 & 13.84 & 12.83 & - & 14.64\\
    & SSIM$\uparrow$  & 0.64 & 0.68 & 0.60 & 0.60 & - & 0.63\\ 
    & LPIPS$\downarrow$ & 0.33 & 0.32 & 0.37 & 0.40 & - & 0.35\\ \hline
\multirow{4}{*}{\shortstack{MaskGaussian~\cite{pruning4}}} 
    & ATE$\downarrow$  & 0.24 & 0.18 & 0.38 & 0.24 & 2.38 & 1.08\\
    & PSNR$\uparrow$  & 20.30 & 21.84 & 19.26 & 18.66 & 17.23 & 19.4\\ 
    & SSIM$\uparrow$  & 0.76 & 0.77 & 0.72 & 0.71 & 0.68 & 0.72\\ 
    & LPIPS$\downarrow$ & 0.29 & 0.28 & 0.32 & 0.33 & 0.35 & 0.31\\ \hline
\multirow{4}{*}{\shortstack{Ours \\ (w/o Tile Budget)}}
    & ATE$\downarrow$  & 0.16 & 0.09 & 0.26 &0.15 & 1.24 & 0.38\\
    & PSNR$\uparrow$  & 24.28 & 25.49 & 23.84 & 22.36 & 20.18 & 23.23\\ 
    & SSIM$\uparrow$  & 0.84 & 0.85 & 0.83 & 0.82 & 0.77 & 0.82\\ 
    & LPIPS$\downarrow$ & 0.21 & 0.20 & 0.23 & 0.24 & 0.29 & 0.23\\ \hline
\multirow{4}{*}{\shortstack{Ours \\ (w/ Tile Budget)}}    
    & ATE$\downarrow$  & \textbf{0.05} & 0.03 & \textbf{0.08} & \textbf{0.05} & 0.13 & { \textcolor{silver}{0.07}}\\
    & PSNR$\uparrow$  & 31.05 & 32.45 & \textbf{30.28} & 30.55 & 28.33 & {\textcolor{silver}{30.53}} \\ 
    & SSIM$\uparrow$  & \textbf{0.98} & \textbf{0.99} & \textbf{0.98} & 0.97 & \textbf{0.96} & { \textcolor{brown}{0.98}}\\
    & LPIPS$\downarrow$ & \textbf{0.05} & \textbf{0.03} & \textbf{0.05} & \textbf{0.05} & 0.08 & {\textcolor{brown}{0.05}}\\ 
\bottomrule
\end{tabular}}
\vspace{-3mm}
\end{table}

\begin{table}[t]
\setlength{\tabcolsep}{0.8mm}
\vspace{4mm}
\renewcommand\arraystretch{1.1}
\caption{Performance on EuRoC~\cite{EUROC}, reported as Peak Memory $\downarrow$ [GB] and FPS $\uparrow$ (lower is better for memory, higher is better for FPS).}
\vspace{-2mm} 
\label{tab:performance-euroc}
\hspace*{0cm}\makebox[\linewidth][c]{
\begin{tabular}{cccccccc}
\toprule
Method                   &Metric& MH01 & MH02 & MH03 & MH04 & MH05 & Avg. \\ \hline
\multirow{2}{*}{LSG-SLAM~\cite{LGSSLAM}}
    & Memory$\downarrow$  & 24.2 & 21.4 & 25.6 & 30.2 & 25.6 & 25.4 \\ 
    & FPS$\uparrow$  & 1.2 & 1.5 & 1.2 & 1.1 & 1.3 & 1.3 \\  \hline
\multirow{2}{*}{\shortstack{MaskGaussian~\cite{pruning4}}} 
    & Memory$\downarrow$  & 16.3 & 13.8 & 17.9 & 22.2 & 17.8 & 17.6\\
    & FPS$\uparrow$  & 1.5 & 1.9 & 1.6 & 1.5 & 1.6 & 1.6\\ \hline
\multirow{2}{*}{\shortstack{Ours}}    
    & Memory$\downarrow$  & \textbf{9.6} & \textbf{8.4} & \textbf{10.2} & \textbf{12.1} & \textbf{10.2} & { \textcolor{brown}{10.1}}\\
    & FPS$\uparrow$  & \textbf{3.3} & \textbf{4.2} & \textbf{3.5} & \textbf{3.3 }& \textbf{3.6} & {\textcolor{brown}{3.6}} \\ 
\bottomrule
\end{tabular}}
\vspace{-3mm}
\end{table}

\begin{table*}[t!]
\setlength{\tabcolsep}{2.4mm}
\renewcommand\arraystretch{1.1}
\caption{Camera tracking and rendering results on KITTI~\cite{KITTI}. Tracking is evaluated by ATE RMSE $\downarrow$ [m], Rendering is evaluated by PSNR $\uparrow$ [dB], SSIM $\uparrow$ [0-1], and LPIPS $\downarrow$ [0-1].  "-" means tracking lost.The results of sequence 01 are not reported, as all methods perform poorly on it.}
\vspace{-2mm} 
\label{tab:accuracy-kitti}
\hspace*{0cm}\makebox[\linewidth][c]{
\begin{tabular}{ccccccccccccc}
\toprule
Method              & \multicolumn{1}{l}{Metrics} & 00    & 02    & 03    & 04    & 05    & 06    & 07    & 08    & 09    & 10    & Avg. \\ \hline
\multirow{4}{*}{LSG-SLAM~\cite{LGSSLAM}} 
& ATE$\downarrow$ & 3.15 & \textbf{9.22} & \textbf{1.88} & 1.62 & \textbf{1.97} & 2.62 & \textbf{1.52} & 8.43 & \textbf{6.06} & \textbf{3.05} & {\textcolor{silver}{3.85}} \\
& PSNR$\uparrow$    & \textbf{27.09} & \textbf{26.64} & 26.18 & \textbf{25.46} & 26.90 & \textbf{27.79} & 25.98 & \textbf{26.17} & \textbf{26.81} & \textbf{26.82} & {\textcolor{brown}{26.58}} \\ 
& SSIM$\uparrow$    & \textbf{0.97} & \textbf{0.97} & \textbf{0.97} & \textbf{0.95} & \textbf{0.97} & \textbf{0.97} & \textbf{0.96} & 0.96 & \textbf{0.97} & \textbf{0.97}& {\textcolor{brown}{0.97}} \\  
& LPIPS$\downarrow$ & \textbf{0.07} & \textbf{0.07} & \textbf{0.07} & \textbf{0.09} & \textbf{0.07} & \textbf{0.06} & \textbf{0.07} & 0.08 & \textbf{0.07} & \textbf{0.07} & {\textcolor{brown}{0.07}} \\ \hline
\multirow{4}{*}{LightGaussian~\cite{pruning1}} 
& ATE$\downarrow$ & - & - & 15.23 & 16.84 & 19.85 & 20.21 & 19.94 & - & - & 21.23 & 18.88 \\
& PSNR$\uparrow$    & - & - & 15.83 & 15.27 & 16.02 & 16.20 & 15.28 & - & - & 16.43 & 15.83 \\  
& SSIM$\uparrow$    & - & - & 0.79 & 0.78 & 0.80 & 0.80 & 0.80 & - & - & 0.79 & 0.79 \\ 
& LPIPS$\downarrow$ & - & - & 0.33 & 0.31 & 0.33 & 0.32 & 0.32 & - & - & 0.32 & 0.32 \\ \hline
\multirow{4}{*}{LP-3DGS~\cite{pruning3}} 
& ATE$\downarrow$ & 9.20 & - & 8.98 & 7.26 & 10.83 & 11.29 & 9.23 & - & - & 12.45 & 9.89 \\
& PSNR$\uparrow$    & 18.57 & - & 18.34 & 17.39 & 18.84 & 19.22 & 18.43 & - & - & 17.76 & 18.36 \\  
& SSIM$\uparrow$    & 0.84 & - & 0.83 & 0.81 & 0.83 & 0.83 & 0.82 & - & - & 0.82 & 0.83 \\ 
& LPIPS$\downarrow$ & 0.24 & - & 0.26 & 0.24 & 0.26 & 0.25 & 0.25 & - & - & 0.24 & 0.25 \\ \hline
\multirow{4}{*}{MaskGaussian~\cite{pruning4}} 
& ATE$\downarrow$ & 5.42 & 14.23 & 4.29 & 3.88 & 4.29 & 5.32 & 4.53 & 13.87 & 12.32 & 6.94 & 7.51 \\
& PSNR$\uparrow$    & 21.72 & 20.83 & 20.98 & 19.29 & 20.31 & 21.97 & 20.57 & 19.99 & 20.03 & 19.98 & 20.56 \\  
& SSIM$\uparrow$    & 0.92 & 0.91 & 0.92 & 0.90 & 0.90 & 0.91 & 0.91 & 0.90 & 0.89 & 0.92 & 0.91 \\ 
& LPIPS$\downarrow$ & 0.18 & 0.20 & 0.18 & 0.22 & 0.19 & 0.18 & 0.18 & 0.16 & 0.19 & 0.19 & 0.19 \\ \hline
\multirow{4}{*}{\shortstack{Ours \\ (w/o Tile Budget)}} 
& ATE$\downarrow$ & 4.23 & 12.48 & 2.47 & 2.19 & 2.95 & 3.34 & 3.26 & 10.63 & 7.62 & 4.56 & 5.37 \\
& PSNR$\uparrow$    & 24.63 & 23.81 & 23.54 & 22.83 & 23.61 & 24.62 & 23.84 & 22.45 & 23.91 & 22.29 & 23.55 \\  
& SSIM$\uparrow$    & 0.95 & 0.94 & 0.95 & 0.93 & 0.93 & 0.94 & 0.94 & 0.93 & 0.92 & 0.95 & 0.94 \\ 
& LPIPS$\downarrow$ & 0.12 & 0.14 & 0.12 & 0.16 & 0.13 & 0.12 & 0.12 & 0.11 & 0.13 & 0.13 & 0.13 \\ \hline
\multirow{4}{*}{\shortstack{Ours \\ (w/ Tile Budget)}}  
& ATE$\downarrow$ & \textbf{3.13} & 9.24 & 1.92 & \textbf{1.58} & \textbf{1.97} & \textbf{2.58} & 1.55 & \textbf{7.32} & 6.83 & 3.22 & {\textcolor{brown}{3.81}} \\
& PSNR$\uparrow$    & 27.01 & 26.31 & \textbf{26.49} & 25.31 & \textbf{26.95} & 27.33 & \textbf{26.31} & 26.10 & 26.33 & 26.55 & {\textcolor{silver}{26.46}} \\ 
& SSIM$\uparrow$    & \textbf{0.97} & \textbf{0.97} & \textbf{0.97} & \textbf{0.95} & \textbf{0.97} & \textbf{0.97} & \textbf{0.96} & \textbf{0.97} & 0.96 & \textbf{0.97} & {\textcolor{brown}{0.97}} \\ 
& LPIPS$\downarrow$ & \textbf{0.07} & \textbf{0.07} & \textbf{0.07} & \textbf{0.09}& \textbf{0.07} & \textbf{0.06} & \textbf{0.07} &\textbf{ 0.07} & 0.08 & 0.08 & {\textcolor{brown}{0.07}} \\
\bottomrule
\end{tabular}}
\vspace{-4mm}
\end{table*}

\begin{table*}[t!]
\vspace{4mm}
\setlength{\tabcolsep}{2.4mm}
\renewcommand\arraystretch{1.1}
\caption{Performance on KITTI~\cite{KITTI}, reported as Peak Memory $\downarrow$ [GB] and FPS $\uparrow$ (lower is better for memory, higher is better for FPS).}
\vspace{-2mm} 
\label{tab:performace-kitti}
\hspace*{0cm}\makebox[\linewidth][c]{
\begin{tabular}{ccccccccccccc}
\toprule
Method              & \multicolumn{1}{l}{Metrics} & 00    & 02    & 03    & 04    & 05    & 06    & 07    & 08    & 09    & 10    & Avg. \\ \hline
\multirow{2}{*}{LSG-SLAM~\cite{LGSSLAM}} 
& Memory $\downarrow$ & 40.2 & 36.6 & 32.1 & 37.1 & 30.6 & 31.2 & 38.5 & 36.2 & 30.4 & 34.2 & 34.7 \\
& FPS $\uparrow$      & 0.7 & 0.8 & 0.9 & 0.7 & 1.1 & 0.9 & 0.7 & 0.8 & 1.0 & 0.8 & 0.8 \\  \hline
\multirow{2}{*}{MaskGaussian~\cite{pruning4}} 
& Memory $\downarrow$ & 35.2 & 31.6 & 26.3 & 32.3 & 24.8 & 25.2 & 32.4 & 30.5 & 24.3 & 29.6 & 29.2 \\
& FPS $\uparrow$      & 0.9 & 1.0 & 1.1 & 0.9 & 1.4 & 1.1 & 0.9 & 1.0 & 1.3 & 1.1 & 1.1 \\   \hline
\multirow{2}{*}{Ours}  
& Memory $\downarrow$ & \textbf{13.9} & \textbf{12.7} & \textbf{10.8} & \textbf{12.9} & \textbf{10.2} & \textbf{10.5} & \textbf{14.0} & \textbf{12.4} & \textbf{10.1 }& \textbf{13.3} & {\textcolor{brown}{11.9}} \\
& FPS $\uparrow$      & \textbf{2.0} & \textbf{2.2} & \textbf{2.5} & \textbf{2.0} & \textbf{3.1} & \textbf{2.4 }& \textbf{1.8} & \textbf{2.3} & \textbf{2.9} & \textbf{2.3 }& {\textcolor{brown}{2.4}}\\ 
\bottomrule
\end{tabular}}
\vspace{-4mm}
\end{table*}

\textbf{Peak Memory and Runtime Performance.} We compare three algorithms that successfully complete all five EuRoC sequences as baselines. 
As shown in Tab.~\ref{tab:performance-euroc}, our method reduces peak memory consumption by 61.3\% on average across the five sequences and achieves a 2.7$\times$ improvement in FPS. 
In contrast, MaskGaussian~\cite{pruning4}, although adopting the same pruning ratio as ours, employs a mask-based strategy in which low-importance Gaussians are not immediately removed but retained in memory for a period of time. 
This deferred deletion results in much weaker memory reduction, yielding only about 30\% savings compared to the baseline. 
Moreover, by substantially lowering peak memory consumption, our method significantly improves runtime efficiency, achieving a 2.2$\times$ speedup over MaskGaussian.

\subsection{Evaluation on KITTI}
\textbf{Tracking and Mapping Accuracy.}
Detailed evaluation results are presented in Tab.~\ref{tab:accuracy-kitti}. 
Our rendering-area--aware pruning strategy consistently achieves tracking accuracy comparable to LSG-SLAM~\cite{LGSSLAM} across all sequences. 
Moreover, in challenging cases such as sequences 00, 04, 05, 06, and 08, by effectively eliminating redundant Gaussians, our method even surpasses LSG-SLAM, demonstrating that carefully designed pruning can not only preserve but also enhance robustness in large-scale outdoor environments. 
By contrast, since the KITTI dataset~\cite{KITTI} targets autonomous driving and its scenes are larger and more expansive than EuRoC~\cite{EUROC}, pruning methods based on gradient or opacity, including LightGaussian~\cite{pruning1} and LP-3DGS~\cite{pruning3}, experience severe tracking loss due to overly  Gaussian removal.

\textbf{Peak Memory and Runtime Performance.}
Detailed evaluation results are presented in Tab.~\ref{tab:performace-kitti}. 
We compare three algorithms that successfully track all KITTI 00--10 sequences: LSG-SLAM~\cite{LGSSLAM}, MaskGaussian~\cite{pruning4}, and our Pocket-SLAM. 
As shown in the table, Pocket-SLAM reduces peak memory consumption by 65.7\% while achieving a 2.9$\times$ speedup in FPS. 
Notably, the improvement on KITTI is even more pronounced than on EuRoC, not only in runtime performance but also in tracking accuracy. 
This is because KITTI, as an autonomous driving benchmark, contains much larger outdoor environments where most of the important Gaussians are concentrated in wide rendering areas such as roads and sky. 
Our rendering-area--aware pruning strategy can more effectively distinguish essential from redundant Gaussians in such settings, yielding substantial memory savings and improved accuracy. 
A more detailed explanation of this mechanism is provided in Sec.~\ref{sec:profiling}.

\begin{figure*}[t!]
    \centering
    \includegraphics[width=0.9\linewidth]{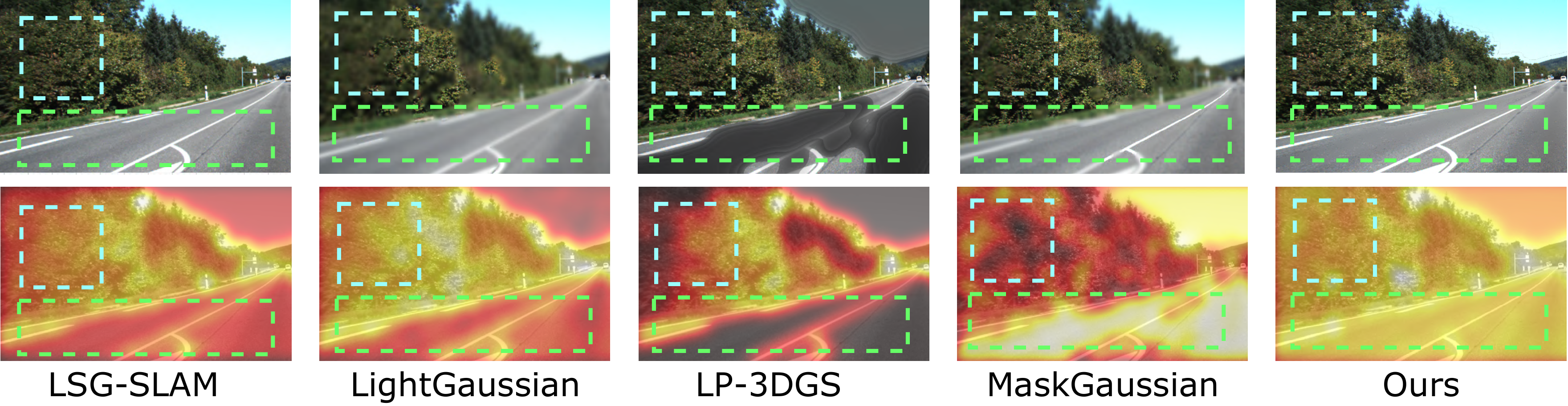}
    \caption{Results on KITTI~\cite{KITTI} sequence 10. We compare five methods: LSG-SLAM~\cite{LGSSLAM}, LightGaussian~\cite{pruning1}, LP-3DGS~\cite{pruning3}, MaskGaussian~\cite{pruning4}, and our approach. For each method, the top row shows the rendered image, while the bottom row presents the Gaussian density heatmap, where brighter colors indicate regions with higher Gaussian counts. Texture-dense and texture-sparse regions are highlighted with blue and green boxes, respectively.}
    \label{fig:accuracy_profiling}
    \vspace{-5mm}
\end{figure*}

\subsection{Result Profiling}
\label{sec:profiling}

\textbf{Accuracy Result Analysis.} Detailed evaluation results are presented in Fig.~\ref{fig:accuracy_profiling}, highlighting the limitations of existing pruning strategies in 3DGS-SLAM. LightGaussian~\cite{pruning1} prunes Gaussians based on opacity, whereas LP-3DGS~\cite{pruning3} relies on gradient magnitude. Both criteria are problematic in outdoor scenarios, since Gaussians covering large but critical regions such as sky and road often exhibit low opacity or gradients and are therefore aggressively removed. This eliminates wide-coverage Gaussians that provide global photometric constraints, leading to unstable pose estimation and frequent drift. In contrast, Pocket-SLAM evaluates Gaussian importance by rendering area, which more accurately reflects their contribution to the image and is particularly well suited for outdoor SLAM and autonomous driving tasks.

MaskGaussian~\cite{pruning4} introduces rendering area as a measure but applies it in a globally area-driven manner, thereby overlooking texture-dense regions. Consequently, fine-grained Gaussians encoding trees, bushes, and façades are masked out, weakening the local consistency on which SLAM losses directly rely. Pocket-SLAM addresses this limitation by augmenting rendering-area pruning with a tile-level budget mechanism, which preserves small Gaussians in texture-rich areas while still pruning redundant ones. This balanced design maintains both global constraints and local detail, enabling significant memory reduction without compromising tracking or reconstruction accuracy.

In summary, Pocket-SLAM addresses limitations of pruning methods by introducing Rendering-Area--Aware Pruning, which resolves the mismatch in Gaussian importance estimation for outdoor scenarios. Additionally, the Tile-Level Budget mechanism mitigates information loss in texture-dense regions caused by purely area-driven pruning. With these designs, Pocket-SLAM achieves stable tracking and accurate mapping while substantially reducing memory consumption.

\begin{figure}[t]
    \vspace{4mm}
    \centering
    \includegraphics[width=1\linewidth]{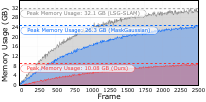}
    \caption{Results on KITTI~\cite{KITTI} sequence 3, showing the trend of memory usage growth with frames for LSG-SLAM~\cite{LGSSLAM}, MaskGaussian~\cite{pruning4}, and Pocket-SLAM. All Gaussian-related parameters are included.}
    \label{fig:memory}
    \vspace{-5mm}
\end{figure}

\textbf{Peak Memory Usage Result Analysis.} We observe that memory consumption fluctuates at each keyframe, since mapping operations periodically introduce new Gaussians. Consequently, the change in the number of Gaussians at each keyframe directly determines the peak memory consumption. Although MaskGaussian~\cite{pruning4} adopts the same pruning ratio as our method, it relies on a mask-pruning strategy in which low-importance Gaussians are temporarily deactivated rather than immediately removed. As a result, a large number of redundant Gaussians remain stored in memory for longer throughout the process. In practice, this means that at each keyframe MaskGaussian removes far fewer Gaussians than our method, which directly eliminates redundant splats. Consequently, its peak memory consumption is reduced only marginally, whereas our approach achieves substantially larger memory savings. This observation is consistent with the trends shown in Fig.~\ref{fig:memory}.

\section{Conclusions and Future Work}
In this paper, we present Pocket-SLAM, a memory-efficient 3DGS-SLAM framework that integrates a \emph{rendering-area--aware pruning strategy} with a \emph{tile-level budget mechanism}. While rendering-area--aware pruning reduces redundant Gaussians by evaluating their pixel coverage, it inevitably removes small Gaussians in texture-dense regions, leading to local information loss and degraded mapping quality. To address this issue, Pocket-SLAM incorporates the tile-level budget mechanism, which allocates survival budgets across tiles according to texture distribution. This ensures that both large-area Gaussians providing global constraints (e.g., roads and sky) and small Gaussians in texture-rich regions (e.g., trees, bushes, façades) are preserved. By combining these two complementary strategies, Pocket-SLAM achieves over 60\% memory reduction and more than $2\times$ FPS improvement on the EuRoC~\cite{EUROC} and KITTI~\cite{KITTI} datasets, while maintaining robust tracking and high-fidelity reconstruction.

Future directions include designing adaptive pruning schedules that dynamically adjust pruning ratios based on scene complexity and temporal variations; exploring long-term, dynamic outdoor scenarios such as autonomous driving at night, across seasonal changes, or under severe weather conditions; and deploying Pocket-SLAM on resource-constrained hardware platforms such as embedded GPUs~\cite{ONX} and edge accelerators~\cite{GauSPU, gsarch, distwar} to enable practical and reliable robotic applications.


\footnotesize

\clearpage

\bibliographystyle{IEEEtran} 
\bibliography{icra_abrv}

\end{document}